%% file: template.tex
\documentclass{article}

\usepackage{arxiv}

\usepackage{amsmath}
\usepackage[utf8]{inputenc} 
\usepackage[T1]{fontenc}    
\usepackage{hyperref}       
\usepackage{url}            
\usepackage{amsfonts}       
\usepackage{nicefrac}       
\usepackage{microtype}      
\usepackage{cleveref}       
\usepackage{lipsum}         
\usepackage{graphicx}
\usepackage{natbib}
\usepackage{doi}

\usepackage{xcolor}
\usepackage{graphicx}
\definecolor{cvprblue}{rgb}{0.21,0.49,0.74}
\usepackage{multirow}
\usepackage{colortbl} 
\usepackage[normalem]{ulem}
\useunder{\uline}{\ul}{}
\usepackage{arydshln}
\usepackage{booktabs}
\usepackage{pifont}
\newcommand{\cmark}{\ding{51}}%
\newcommand{\xmark}{\ding{55}}%
\usepackage{wrapfig}
\usepackage{algorithmic}
\usepackage{algorithm}
\usepackage{makecell}

\renewcommand{\headeright}{}
\renewcommand{\undertitle}{}

\title{DeFakeQ: Enabling Real-Time Deepfake Detection on Edge Devices via Adaptive Bidirectional Quantization}

\date{}

\newif\ifuniqueAffiliation
\uniqueAffiliationtrue

\ifuniqueAffiliation 
\author{
Xiangyu Li\thanks{Equal contribution} \\
Digital Trust Centre\\
Nanyang Technological University\\
\texttt{xinagyu.li@ntu.edu.sg} \\
\And
Yujing Sun$^*$\thanks{Corresponding author} \\
Digital Trust Centre\\
Nanyang Technological University\\
\texttt{yujing.sun@ntu.edu.sg} \\
\And
Yuhang Zheng \\
Digital Trust Centre\\
Nanyang Technological University\\
\texttt{yuhang.zheng@ntu.edu.sg} \\
\And
Yuexin Ma \\
School of Information Science and Technology\\
ShanghaiTech University\\
\texttt{mayuexin@shanghaitech.edu.cn} \\
\And
Kwok-Yan Lam \\
Digital Trust Centre\\
Nanyang Technological University\\
\texttt{kwokyan.lam@ntu.edu.sg} \\
}

\begin{document}
\maketitle

\input{sec/0_abstract}
\input{sec/1_introduction}
\input{sec/2_related}
\input{sec/3_method}
\input{sec/4_experiment}
\input{sec/5_conclusion}

\bibliographystyle{unsrtnat}
\bibliography{references}  






\end{document}

%% file: sec/0_abstract.tex
\begin{abstract}
Deepfake detection has become a fundamental component of modern media forensics. Despite significant progress in detection accuracy, most existing methods remain computationally intensive and parameter-heavy, limiting their deployment on resource-constrained edge devices that require real-time, on-site inference. This limitation is particularly critical in an era where mobile devices are extensively used for media-centric applications, including online payments, virtual meetings, and social networking.
Meanwhile, due to the unique requirement of capturing extremely subtle forgery artifacts for deepfake detection, state-of-the-art quantization techniques usually underperform for such a challenging task. These fine-grained cues are highly sensitive to model compression and can be easily degraded during quantization, leading to noticeable performance drops. This challenge highlights the need for quantization strategies specifically designed to preserve the discriminative features essential for reliable deepfake detection.
To address this gap, we propose \emph{DefakeQ}, the first quantization framework tailored for deepfake detectors, enabling real-time deployment on edge devices. Our approach introduces a novel adaptive bidirectional compression strategy that simultaneously leverages feature correlations and eliminates redundancy, achieving an effective balance between model compactness and detection performance.
Extensive experiments across five benchmark datasets and eleven state-of-the-art backbone detectors demonstrate that \emph{DeFakeQ} consistently surpasses existing quantization and model compression baselines. Notably, it reduces model size to 10–20\% of the original while retaining up to 90\% of baseline accuracy, substantially outperforming prior state-of-the-art methods.
Furthermore, we deploy \emph{DefakeQ} on mobile devices in real-world scenarios, demonstrating its capability for real-time deepfake detection and its practical applicability in edge environments.

\end{abstract}

\keywords{Deepfake Detection \and Edge Devices \and Post-Training Quantization \and Real-time and Real-world Deployment}

%% file: sec/1_introduction.tex
\section{Introduction}
\label{sec:intro}

Deepfake detection has become increasingly important with the rapid evolution of generative artificial intelligence, forming a crucial foundation for media forensics~\cite{chu2015information}.
We notice that current mainstream approaches~\cite{huang2025sida,fu2025exploring,tan2025c2p, wang2026map, yang2023masked} mainly focus on improving detection accuracy but often rely on relatively high computational resources~\cite{wang2025idcnet, xia2024mmnet, zhao2023istvt}. 
However, such resources are not readily available in everyday settings, where most users rely on edge devices such as mobile phones for payment, work, and entertainment, contexts that may involve AI-generated content posing potential risks or fraud, requiring real-time, on-site Deepfake identification.

\begin{figure}[t]
	\centering
	\includegraphics[width=0.8\columnwidth]{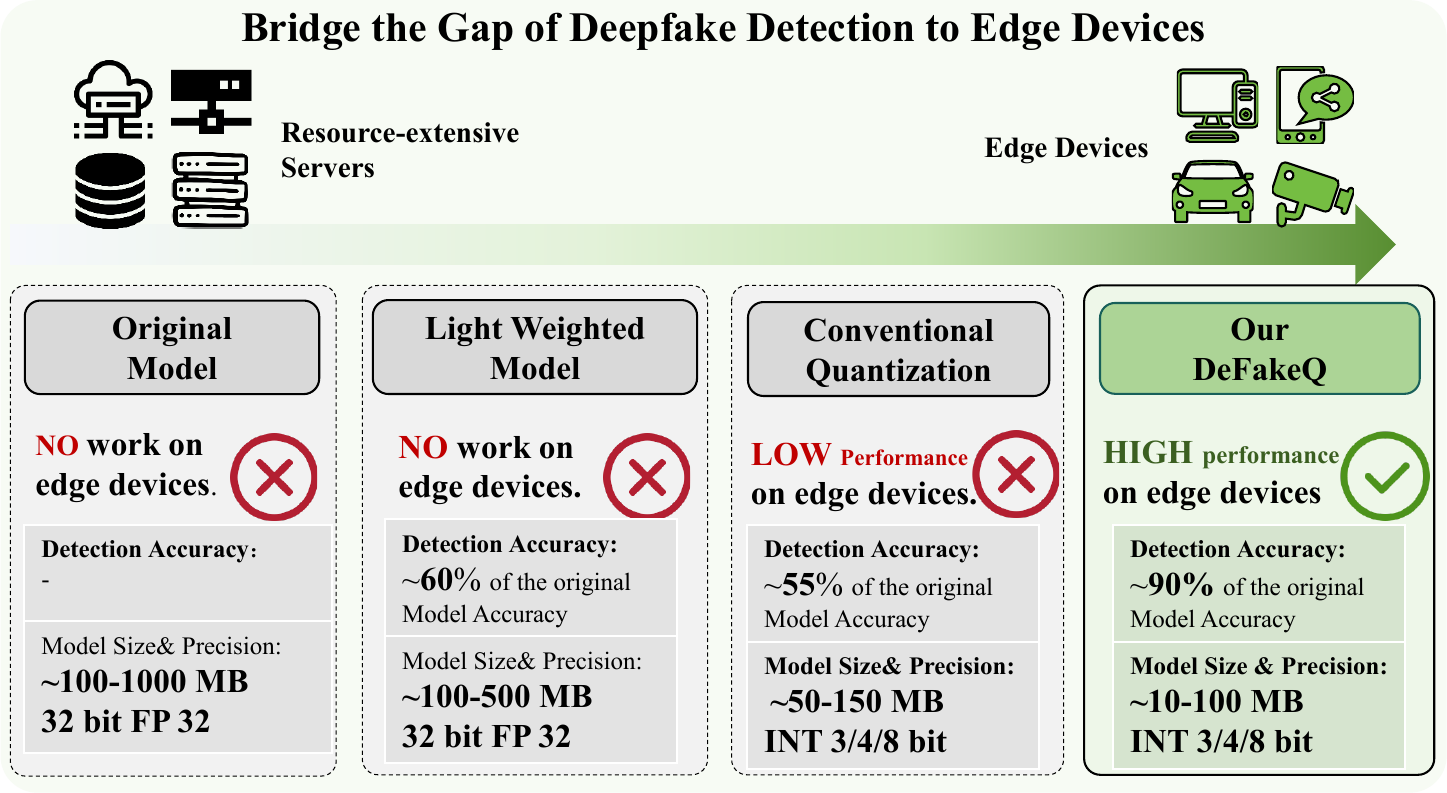}
	\caption{We present the first quantization framework tailored for deepfake detection, \emph{DeFakeQ}, enabling real-time deployment on edge devices. Compared to existing lightweight model designs and conventional quantization approaches, our framework significantly reduces computational cost while maintaining high detection accuracy.}
	\label{fig_motivation}
\end{figure}

Nevertheless, edge devices operate under strict constraints, including limited memory, restricted computational power, and stringent latency requirements. 
For instance, a state-of-the-art Xception-based deepfake detector~\cite{tan2019efficientnet} requires \emph{28M} parameters and \emph{1.2G} FLOPs per inference, resulting in a \emph{350ms} latency even on a high-end mobile System-on-Chip, well above the threshold for real-time applications. Therefore, it is crucial to develop efficient adaptation strategies that enable existing deepfake detectors to run effectively on edge devices.

Realizing the importance of bridging the edge deployment gap for state-of-the-art deepfake detectors, researchers have explored model pruning~\cite{vidya2023compressed, kim2021fretal, li2023improvement} and lightweight architecture design~\cite{wang2023face, dutta2024enhancing}. However, both approaches share a key limitation. Whether reducing redundant parameters or designing compact networks, they still depend on full-precision weight representations. This reliance results in substantial memory overhead, as even pruned or lightweight models retain large numbers of full-precision weights to maintain feature extraction capacity, making them unsuitable for memory-constrained edge devices.

To address this limitation, Post-Training Quantization (PTQ) appears to be an effective backbone for enabling efficient deepfake detection on edge devices. By compressing model weights and activations from full-precision to low-bit representations, PTQ substantially reduces memory consumption and computational demand.

However, directly applying existing general-purpose or task-specific quantization methods~\cite{nagel2021white} to deepfake detection remains ineffective, largely due to the task’s unique nature. Deepfake detection fundamentally relies on capturing ultra-subtle, fine-grained facial forgery cues, such as imperceptible micro-expression inconsistencies and faint texture anomalies across critical facial regions. These cues are easily disrupted when model weights and activations are compressed from floating-point to low-precision integer formats, as quantization inevitably introduces noise and information loss. Consequently, even minor distortions can obscure the delicate features essential for reliable detection, underscoring the urgent need for quantization strategies tailored specifically to the sensitivity and complexity of deepfake detection.

To enable deepfake detection on edge devices, for the first time, we propose \textbf{\emph{DeFakeQ}}, an adaptive bidirectional quantization framework designed to balance model efficiency and accuracy. \emph{DeFakeQ} tackles the core challenge of edge deployment: achieving lightweight, real-time inference while preserving the fine-grained facial forgery cues essential for reliable detection.
\emph{DeFakeQ} incorporates two key innovations to overcome the limitations of existing quantization methods. First, \textbf{\emph{Horizontal Adaptive Block Quantization (HAQ)}} adaptively adjusts the quantization bit-width across each layer within the block by calculating the importance of weights and activations for each layer. It assigns higher bit-widths to layers with high information density to preserve their feature representation capability, thereby minimizing information loss while maximizing computational and storage efficiency. Subsequently, \textbf{\emph{Vertical Efficient Feature Fine-Tuning (VEFT)}} randomly selects a small subset of feature channels and restores them to their original full-precision format. By constructing a progressive contrastive metric learning loss function, VEFT further preserves discriminative features within quantization blocks. This complements HAQ’s layer-level optimization and further enhances the model’s accuracy in resource-constrained scenarios.


Extensive experiments on 5 Deepfake datasets across 7 mainstream SOTA Deepfake detectors demonstrate that \emph{DeFakeQ} reduces pre-trained models’ parameter size to merely 10\% of their full-precision counterparts while retaining over 90\% of the original full-precision performance. Moreover, we have successfully deployed \emph{DeFakeQ} on mobile devices for real-time and real-world applications, directly validating its practical real-world application value for resource-constrained edge deployment scenarios.

In summary, our contributions are as follows:
\begin{itemize}
\item We are the first to introduce \textbf{\emph{DeFakeQ}}, an Adaptive Bidirectional Quantization framework tailored for deployment of real-time deepfake detection on edge devices.

\item In \emph{DeFakeQ}, we novelly propose \textbf{\emph{Horizontal Adaptive Block Quantization (HAQ)}} to dynamically allocate bit-width based on the information saliency of each block and \textbf{\emph{Vertical Efficient Feature Fine-Tuning (VEFT)}} to maximize the preservation of discriminative features within quantized blocks, achieving a better trade-off between detection accuracy and compression efficiency.
\item Extensive experiments on benchmark Deepfake datasets across 11 mainstream SOTA / Backbone Deepfake detectors demonstrate that \emph{DeFakeQ} reduces pre-trained models' parameter size to merely 10\% of their full-precision counterparts while retaining over 90\% of the original full-precision performance. Moreover, we have successfully deployed and worked \emph{DeFakeQ} on actual mobile devices, directly validating its practical real-world application value for resource-constrained edge deployment scenarios.
\end{itemize}

%% file: sec/2_related.tex
\section{Related Works}
\label{sec:related}

\subsection{Deepfake Detection Models}
The most recent deepfake detection focuses on improving generalization across different domains and forgery types. For image forgery detection, key approaches include data augmentation~\cite{chen2022self, li2020face, li2018exposing, shiohara2022detecting, zhao2021learning, yang2021mtd}, frequency-domain clue extraction~\cite{gu2022exploiting, liu2021spatial, luo2021generalizing, qian2020thinking, wang2023dynamic}, Identity Information (ID) utilization~\cite{dong2023implicit, huang2023implicit}, disentanglement learning~\cite{liang2022exploring, yan2023ucf, yang2021learning}, custom network design~\cite{zhao2021multi, dang2020detection}, reconstruction learning~\cite{cao2022end, wang2021representative, wang2026fauforensics}, and 3D decomposition~\cite{zhu2021face}. Recently, studies~\cite{jia2024can, tan2024data, zheng2026boosting} have focused on enhancing generalization through specific training-free pipelines. However, these models typically involve large parameter sizes and high computational costs, making them incompatible with resource-constrained edge devices. This incompatibility significantly hinders their practical deployment in real-world scenarios, thereby limiting their actual application value. Although researchers have explored model pruning~\cite{vidya2023compressed, kim2021fretal} and lightweight architecture design~\cite{wang2023face, dutta2024enhancing} to tackle the deployment of deepfake detectors, their full-precision representation still incurs substantial storage overhead. Meanwhile, model compression induced by pruning inevitably adversely affects detection performance.

\subsection{Model Quantization}
Model quantization~\cite{nagel2021white} tackles deployment on resource-constrained devices via converting high-precision floats to low-bit integers. While Quantization-Aware Training (QAT)~\cite{lee2021network, li2022q, zhu2023quantized} delivers high accuracy, full-dataset retraining makes it prohibitively expensive. Post-Training Quantization (PTQ)~\cite{frantar2022gptq, li2023repq, dettmers2023qlora, huang2024billm} is more practical, relying on a small unlabeled calibration set and no retraining. For instance, AdaRound~\cite{nagel2020up}, which introduces a learnable task-aware rounding mechanism; BRECQ~\cite{librecq}, which employs blockwise reconstruction and leverages Fisher information to protect critical weights; QDrop~\cite{weiqdrop}, which adopts Dropout-inspired regularization to enhance resilience against quantization noise; and PD-Quant~\cite{liu2023pd}, which utilizes global information (e.g., discrepancies in final model predictions). However, these methods are not adapted to deepfake detection, which rely on capturing subtle facial variations. Existing quantization methods inevitably lose fine-grained cues during the quantization, failing to preserve the discriminative details essential for edge deepfake detection.

%% file: sec/3_method.tex
\begin{figure*}[t]
	\centering
	\includegraphics[width=1\textwidth]{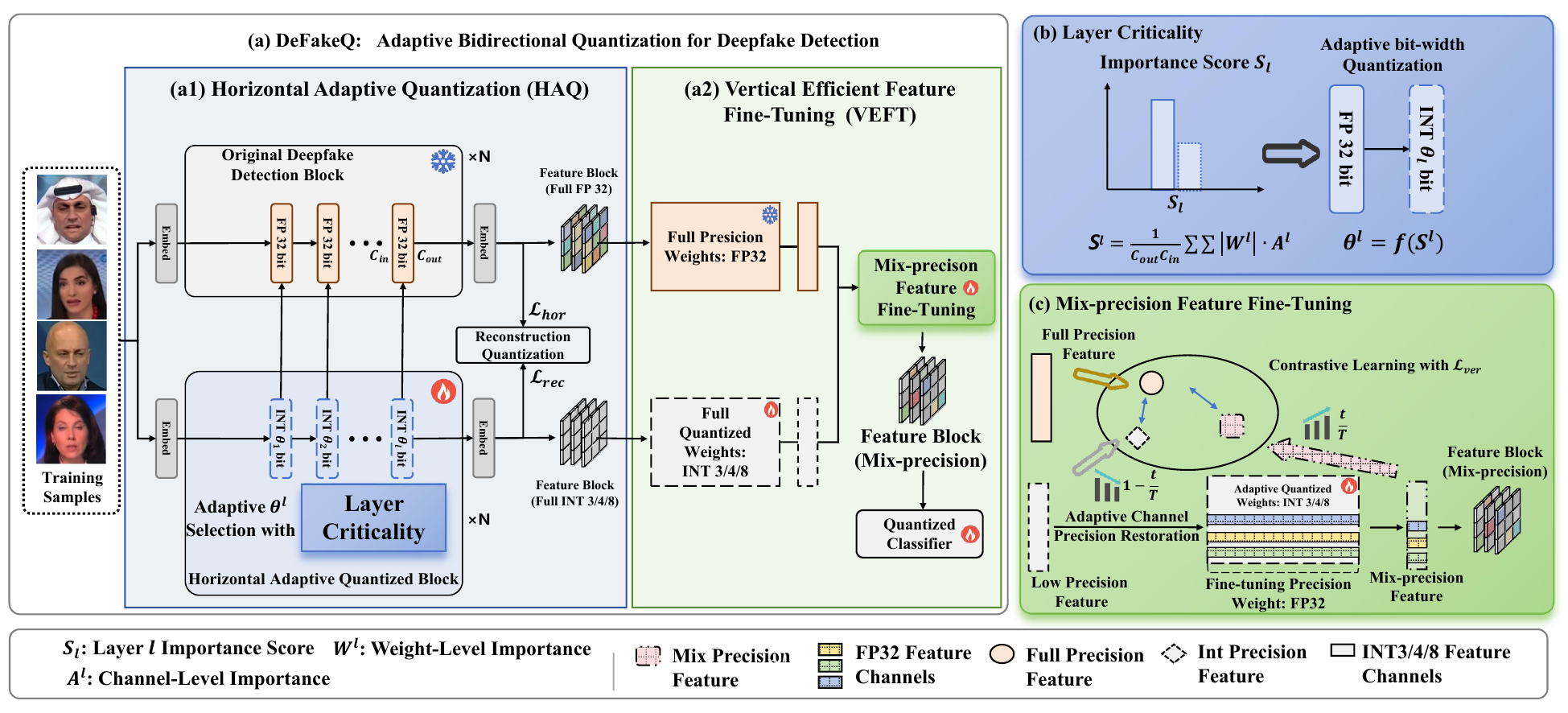}
	\caption{The overview of \emph{DeFakeQ}. It comprises two core components: Horizontal Adaptive Quantization (HAQ) and Vertical Efficient Feature Fine-Tuning (VEFT). HAQ first calculates the importance scores of weights within each network block. Guided by a specially designed horizontal loss function $\mathcal{L}_{hor}$, it adaptively identifies blocks rich in discriminative information and assigns them higher bit-widths, ensuring critical forgery-related features are preserved during quantization. Complementing HAQ, VEFT operates at the channel level: it adaptively identifies and prioritizes sensitive feature channels within each quantized block. This mechanism maximizes the retention of discriminative features in quantized blocks while maintaining efficient compression of redundant channels.
    }
	\label{fig1}
\end{figure*}

\section{Background and Preliminaries of Quantization}
Model quantization~\cite{nagel2021white} has emerged as a pivotal technique for accelerating inference and compressing deep learning models, as it enables the reduction of parameter bit-width representations while preserving the integrity of model accuracy. For mainstream quantization paradigms, the operations of quantization and de-quantization are formally defined as follows:
\begin{equation}
\begin{aligned}
\label{eq1}
\texttt{int} = &\texttt{clamp}\left(\left\lfloor\frac{fp}{s}\right\rceil + z, 0, 2^{k} - 1\right), \\ 
&\hat{fp} = s \cdot (int - z) \approx fp,
\end{aligned}
\end{equation}
where $\emph{s}$ and $\emph{z}$ denote the scaling factor and zero point, respectively. $\left\lfloor \cdot \right\rceil$ represents the Round-to-Nearest (RTN) rounding operator. $f_{p}$ corresponds to the original floating-point value, while $\hat{f}_{p}$ stands for the resulting de-quantized value. $\texttt{int}$ denotes the integer value mapped from the floating-point input via quantization, and the $\texttt{clamp}$ function is used to clip any values that lie outside the range of a $k$-bit integer.

\textbf{Why Conventional Quantization ineffective in Deepfake Detection ?} \quad
Traditional quantization methods exhibit limited effectiveness when directly applied to deepfake detection. This task fundamentally depends on capturing fine-grained forgery artifacts, such as micro-expression inconsistencies, boundary artifacts, and subtle texture anomalies. However, the aggressive precision reduction inherent in conventional quantization often leads to the degradation or loss of these discriminative signals, thereby compromising the detector’s reliability and overall performance.
Moreover, such clues are typically sparse, fragile, and unevenly distributed across network blocks~\cite{dong2022explaining, li2025learnable}, with certain layers encoding disproportionately more forensic information than others. Traditional uniform bit-width quantization fails to account for this heterogeneity: applying low-bit quantization (e.g., 4-bit) universally risks severe loss of fragile cues, while high-bit quantization (e.g., 8-bit) across all layers undermines efficiency by over-preserving redundant, non-forensic information.

\section{Method}
\label{sec:method}

The large model size and high computational demands of mainstream deepfake detectors make them impractical for deployment on resource-constrained edge devices.
Although lightweight network design~\cite{chen2022defakehop++} and pruning~\cite{han2015deep} have been explored to reduce complexity, their practical usability on edge platforms remains limited, as they still rely on full-precision (FP16/FP32) parameters, leading to substantial storage overhead.
Meanwhile, conventional quantization approaches~\cite{weiqdrop} also underperform in deepfake detection tasks, failing to preserve the subtle forensic cues essential for reliable classification.

To bridge the gap, we propose the first deepfake detection quantization approach to enable real-time edge device deployment, \emph{DeFakeQ}.  
We significantly reduce both model size and storage requirements through an adaptive bidirectional precision compression scheme 
while maintaining high detection accuracy.

The pipeline of \emph{DeFakeQ} is demonstrated in Fig.~\ref{fig1}(a). In general, we propose an adpative bidirectional strategy to resolve the contradiction between reducing model size and maintaining important forensics clues. Our \emph{DeFakeQ} comprises two key components. To address the limitation of uniform bit-width in conventional quantization where fragile cues are at risk of severe loss, \textbf{\emph{Horizontal Adaptive Quantization (HAQ)}} is proposed. The goal of HAQ is to adaptively select bit-widths while maximizing the approximation to the original floating-point (FP) model. This ensures that blocks rich in discriminative features retain higher precision, enabling efficient compression while preserving critical forgery-related information. In addition, to excavate the feature representation ability within each block, we propose \textbf{\emph{Vertical Efficient Feature Fine-Tuning (VEFT)}} to adaptively select sensitive feature channels within each quantized block, maximizing the retention of discriminative features in quantized blocks while maintaining the compression efficiency of redundant features.

\subsection{Horizontal Adaptive Quantization}
To overcome the limitation of uniform Bit-Width in conventional quantizations, \textbf{\emph{Horizontal Adaptive Quantization (HAQ)}} is thus designed to dynamically allocate bit-width based on the information saliency of each block (as shown in Figure~\ref{fig1}(a1)). 
Generally, blocks rich in forgery-relevant discriminative features are assigned higher bit-widths (e.g., 8-bit) to preserve their forensic integrity. 
While blocks with redundant or non-critical information are quantized to lower bit-widths (e.g., 4-bit) to reduce computational overhead. 
This targeted strategy ensures the most valuable forensic cues remain intact while achieving substantial compression. 

We quantifies the ``\emph{Criticality}'' of each linear layer $l$ via a Weight-Activation Importance Score $S^l$ (as shown in Figure~\ref{fig1}(b)), which integrates both weight magnitude and channel importance energy to capture the layer's true impact on model output.
Specifically, let the input activation tensor of layer $l$ be $X^{(l)} \in \mathbb{R}^{N \times C_{in}}$, where $N$ denotes the number of tokens in the activation tensor (corresponding to the spatial dimensions of feature maps or sequence length in transformers); $C_{in}$ represents the number of input channels (i.e., the feature dimension); $X_{n,j}^{(l)}$ is the activation value of the $n_{th}$ token in the $j_{th}$ input channel.

\textbf{Channel-Level Importance.}\quad
Input activation tensors exhibit inherent variability in information density across channels. To quantify this disparity, we adopt the $L_{2}$ norm (\emph{Energy norm}) as a metric for channel-level importance due to its ability to effectively characterize the \emph{Energy Intensity} of activations: higher $L_{2}$ norm indicates greater information density, reflecting the cumulative contribution of a channel's activations to the layer's output~\cite{devoto2024simple,liu2025patchprune}. Thus, for each input channel $j$, we compute its average $L_{2}$ norm to ensure generalization beyond individual samples. This yields the input channel importance vector $A^{(l)} \in \mathbb{R}^{C_{in}}$, where each element $A_j^{(l)}$ is defined as:
\begin{equation}
\label{eq3}
A_{j}^{(l)} = \left( \frac{1}{M} \sum_{m=1}^{M} \sum_{n=1}^{N} \left( X_{n,j}^{(l,m)} \right)^2 \right)^{\frac{1}{2}} \quad \forall j \in \{1, 2, ..., C_{in}\},
\end{equation}
where $M$ denotes the number of samples in the calibration dataset, and $X_{n,j}^{(l,m)}$ denotes the activation value of the $n_{th}$ token in the $j_{th}$ channel of layer $l$ for the $m$ th calibration sample. The normalization by $M$ ensures that $A^{(l)}$ is robust to variations in the scale of the dataset, while the $L_{2}$ norm directly maps to the channel's informational contribution.

\textbf{Weight-Level Importance.}\quad
Each layer output is determined by the linear transformation of input activations via its weight matrix, which means that the impact of an individual weight element $W_{i,j}^{(l)}$ (connecting the $j_{th}$ input channel to the $i_{th}$ output channel) depends not only on its own magnitude but also on the information density of the input channel it acts upon. A large weight connected to a low-information input channel (low $A_j^{(l)}$) will have minimal influence on the output, whereas a moderate weight linked to a high-information channel can significantly shape the layer's representational output. To capture this synergistic effect, we define an element-wise importance metric with the same dimensions as the weight matrix (where $C_{out}$ is the number of output channels). Each element of this metric is formulated as the product of the absolute weight magnitude (avoid cancelation of positive and negative contributions) and the corresponding input channel importance, which can be formulated as:
\begin{equation}
\label{eq4}
    |W_{i,j}^{(l)}| \cdot A_j^{(l)} \quad \forall i \in \{1, ..., C_{out}\}, \ j \in \{1, ..., C_{in}\}.
\end{equation}
The formulation ensures that directly quantifies the Effective Contribution of each weight-element to the layer's output, effectively filtering out weights that are either small in magnitude or connected to low-information input channels.

\textbf{Weight-Activation Importance Score} $S^l$ for layer $l$ is finalized as the mean of all elements:
\begin{equation}
\label{eq5}
S^l = \frac{1}{C_{out} \cdot C_{in}} \sum_{i=1}^{C_{out}} \sum_{j=1}^{C_{in}} |W_{i,j}^{(l)}| \cdot A_j^{(l)}.
\end{equation}

Thus, the score $S^l$ determines the weight bit-width of the $l$-th layer, where $\theta ^{l} = f(S^l)$ denotes the number of bits assigned to the $l$-th layer and is learned through an optimization process. A higher $S^{l}$ indicates that the layer plays a more critical role in sustaining the model's representational capacity, thereby necessitating higher bit-widths to mitigate substantial accuracy degradation during the quantization process.

\textbf{Learnable Optimization Process.}\quad
The goal of HAQ is to assign a bit-width $b_{l} \in \mathcal{B}$ (a predefined set of candidate bit-widths) to each layer $l \in \mathbf{L}$ (the set of quantizable layers) such that two objectives are balanced: maximizing the accuracy while minimizing the bit-width for each block.
Formally, the HAQ problem is cast as a learnable constrained combinatorial optimization problem:
\begin{equation}
\label{eq6}
    \begin{aligned}
        \min_{b_l \in \mathcal{B}} \quad & \sum_{l \in \mathbf{L}} S^l \cdot \mathcal{E}(b_l), \\
\text{s.t.} \quad & \frac{\sum_{l \in \mathbf{L}} b_l \cdot p_l}{\sum_{l \in \mathbf{L}} p_l} \leq B_{avg},
    \end{aligned}
\end{equation}
where $p_l$ denotes the number of parameters in layer $l$, and $B_{avg}$ denotes the target average bit-width. $\mathcal{E}(b_l)$ denotes a monotonic error function that maps bit-width $b_l$ to the theoretical quantization error of layer $l$. We adopt $\mathcal{E}(b_l) \propto 2^{-b_l}$, a well-validated approximation in quantization~\cite{nagel2020up}, to reflect quantization error roughly increases as bit-width decreases. In Equation~\ref{eq6}, high-$S^l$ layers are allocated more bits to minimize their $\mathcal{E}(b_l)$, while low-$S^l$ layers are assigned fewer bits to save resources, which balances accuracy preservation and efficiency.

As for the HAQ problem, the loss function must be differentiable to enable gradient-based optimization even though bit-widths are inherently discrete. Therefore, the trainable loss function $\mathcal{L}_{hor}$ is defined as:
\begin{equation}
\label{eq7}
    \mathcal{L}_{hor} = \sum_{l \in \mathbf{L}} S^l \cdot \mathcal{E}(\hat{b}_l) + \lambda \cdot \max\left(0, \frac{\sum_{l \in \mathbf{L}} \hat{b}_l \cdot p_l}{\sum_{l \in \mathbf{L}} p_l} - B_{\text{avg}}\right)^2,
\end{equation}
where $\hat{b}_l$ denotes the continuous relaxation of the discrete bit-width $b_l$ and $\lambda$ is penalty coefficient to balance the two terms.

\textbf{Reconstruction Loss Function.}\quad During reconstruction quantization process, $\mathcal{L}_{rec}$ is employed to minimize the discrepancy between the outputs of the quantized and original models. Similar to AdaRound~\cite{nagel2020up}, $\mathcal{L}_{rec}$ is defined as:
\begin{equation}
\label{eq2}
    \mathcal{L}_{rec} = \| \textbf{Wx} - \tilde{\textbf{W}}\textbf{x} \|_F^2 + \alpha R(\textbf{V}),
\end{equation}
where the first term corresponds to the MSE loss; $\textbf{W}$ denotes the weights of the reconstruction layer for original block, $\textbf{x}$ represents its input, $\alpha$ is a trade-off hyperparameter, and $\tilde{\textbf{W}}$ denotes the quantized weights of the quantized block, which can be formulated as:
\begin{equation}
    \tilde{\textbf{W}} = \texttt{clamp}\left( \left\lfloor \frac{\textbf{W}}{s} \right\rfloor + h(\textbf{V}) + z, \, 0, \, 2^k - 1 \right).
\end{equation}

The other term $R(\textbf{V})$, is a differentiable regularizer that encourages the optimization variables $h(\textbf{V}_{i,j})$ to converge to either 0 or 1.

\subsection{Vertical Efficient Feature Fine-Tuning}
The Horizontal Adaptive Quantization (HAQ) has achieved dynamic bit-width allocation at the block level to effectively reduce the computational efficiency. However, HAQ treats each block as a uniform unit, but low-bit blocks still contain a mix of discriminative and redundant features. For example, even a 4-bit block may include discriminative informations sensitive to deepfake detetcion. Uniform 4-bit quantization compresses both discriminative and irrelevant features, leading to the dilution of key forgery artifacts. In addition, even high-bit quantization can cause subtle distortion of these fragile discriminative cues, which are pivotal for distinguishing fakes from reals. 

To bridge this gap, we propose \textbf{\emph{Vertical Efficient Feature Fine-Tuning (VEFT)}}, which maximizes the retention of discriminative features in quantization blocks while maintaining the compression efficiency of redundant features, complementing HAQ’s block-level optimization (as shown in Figure~\ref{fig1}(a2)). For each quantized block, VEFT randomly selects a small subset of feature channels to restore to the original full-precision format, consistent with the precision of the pre-trained full-precision Deepfake detector $M_f$. The remaining channels in each quantized block retain their HAQ-allocated int bit-width, ensuring that redundant information continues to be compressed effectively.  The resulting mix-precision model, denoted as $M_q^v$, is distinguished from HAQ’s low-presicion quantized model $M_q$ (where all quantized blocks retain only int precision).

\textbf{Feature Alignment with contrastive learning.} To align the distribution of $M_q^v$ and $M_q$ with that of $M_f$, VEFT introduces a progressive contrastive learning workflow (as shown in Figure~\ref{fig1}(c)). The anchor sample $x_{a}$ is defined as the output feature of $M_f$ for a given input sample $x$, leveraging $M_f$’s ability to preserve all subtle forgery artifacts as the ground truth for feature distribution. Two positive samples are integrated to guide optimization across all quantized blocks: the first positive sample $x_{pos1}$ is the output feature of $M_q$ (HAQ’s low-presicion model) for the same input, representing the baseline block-level quantization result that requires refinement to reduce discriminative loss; the second positive sample $x_{pos2}$ is the output feature of $M_q^v$, whose partial FP channels make it inherently closer to $x_{a}$ and thus the primary target for alignment. Then, a progressive weight $w(t)$ (where $t$ is the current training epoch and $T$ is the total number of epochs) is introduced to fuse the two positive samples into a combined positive sample $x_{comb} = w(t) \cdot x_{pos2} + (1-w(t)) \cdot x_{pos1}$, with $w(t) = \frac{t}{T}$, ensuring early training prioritizes $x_{pos1}$ to quickly align low-presicion presicion features with $M_f$ and late training emphasizes $x_{pos1}$ to optimize the mix-precision model's accuracy.

\begin{algorithm}[t]
  \caption{\emph{DeFakeQ}: Enabling Real-Time Deepfake Detection on Edge Devices via Adaptive Bidirectional Quantization}
  \label{alg:ptq}
  \textbf{Input}: Pre-trained Full-precision Deepfake Detection Model $M_{f}$, Calibration Set $\emph{D}_{calib}$\\
  \textbf{Output}: Low-presicion quantized model $M_q$, Mix-precision model $M_q^v$
\\
  \textbf{Variables}: The Quantized Layers $\mathbf{L}_{q}$, Scaling Factor $s$, Zero Point $z$
  \begin{algorithmic}[1]
    \STATE Initialize quantized model $M_Q$ by copying parameters from $M_F$ and inserting pseudo-quantization modules
    \STATE Activate observers associated with pseudo-quantization modules to track activation statistics
    \STATE Perform calibration: forward samples from $\emph{D}_{calib}$ into $M_{q}$ through $M_{q}$, observers record layer-wise activation distributions
    \STATE Enable quantization logic of pseudo-quantization modules to simulate low-bit operations
    \FOR{each layer $l \in \mathbf{L}_{q}$ in $M_{q}$}
      \STATE Compute scaling factor $s$ and zero point $z$ for layer $l$ using $\mathcal{L}_{rec}$, $\mathcal{L}_{hor}$ and $\mathcal{L}_{ver}$
    \ENDFOR
    \STATE \textbf{return} Quantized Detection Model $M_{q}$, Mix-precision model $M_q^v$
  \end{algorithmic}
\end{algorithm}

\textbf{Loss Function.}\quad To materialize the progressive contrastive learning workflow above, we propose the loss function $\mathcal{L}_{ver}$ to enforce alignment between the feature distributions of $M_q^v$, $M_q$, and the full-precision ground truth $M_f$, while robustly distinguishing authentic from forged feature patterns. The loss function $\mathcal{L}_{ver}$ is formalized as:
\begin{equation}
\label{eq8}
\mathcal{L}_{ver} = -\log\!\bigg( 
  \frac{e^{\mathrm{sim}(x_a, x_{comb})/\tau}}
       {e^{\mathrm{sim}(x_a, x_{comb})/\tau} + \sum_{k=1}^K e^{\mathrm{sim}(x_a, x_{k})/\tau}}
  \bigg),
\end{equation}
where $\tau$ denotes Temperature coefficient to control the sharpness of the similarity distribution. $K$ denotes Number of negative samples, ensuring diverse negative distributions and robust class discrimination.

The loss function $\mathcal{L}_{ver}$ ensures that the mix-precision model $M_q^v$ not only retains the efficiency of quantized channels but also refines the discriminative power of restored full-precision channels.

\subsection{Total Loss Function of DeFakeQ}
Eventually, the total loss of our proposed \emph{DeFakeQ} can be formulated as:
\begin{equation}
\label{eq9}
    \mathcal{L} = \mathcal{L}_{rec} + \gamma _{1} \mathcal{L}_{hor} + \gamma _{2} \mathcal{L}_{ver},
\end{equation}
where $\gamma _{1}$ and $\gamma _{2}$ are trade-off parameters.

\subsection{Calibrations}
Quantization requires calibration as a critical step to determine optimal quantization parameters (e.g., scaling factors and min/max ranges) for both model weights and activations. Different from the large scale of training data, calibration in quantization requires  small, representative datasets to capture the core distribution of model activations needed to determine optimal quantization parameters, while avoid the excessive computational cost and slower calibration speed of large datasets.

\textbf{Calibration Dataset $\emph{D}_{calib}$.} \quad Considering the cross data evaluation setting where test data is inaccessible during non-test phases, our calibration dataset is constructed entirely from the training dataset (FF+~\cite{rossler2019faceforensics++}). Specifically, we randomly select 256 images to form the calibration dataset, with the selected images deliberately avoiding being sourced from the same video as much as possible.

\textbf{Calibration Process.} \quad Our \emph{DeFakeQ} incorporates pseudo-quantization factors into the full-precision Deepfake detection model (denoted as $M_{f}$) to derive the quantized model (denoted as $M_{q}$). We then activate the observer for these pseudo-quantization factors, where a small calibration dataset $\emph{D}_{calib}$ serves to assess and document the activation distributions across each layer under simulated quantization conditions. Guided by $\emph{D}_{calib}$, the quantized model undergoes layer-by-layer adaptive bit-width reconstruction (via the loss function $\mathcal{L}_{hor}$), with the goal of closely aligning quantized outputs with those of the original floating-point model. After adaptive bit-width reconstruction, we randomly select a small subset of feature channels to restore to the original full-precision format for each quantized block, consistent with the precision of the pre-trained full-precision Deepfake detector $M_{f}$. The remaining channels in each quantized block retain their HAQ-allocated int bit-width, ensuring that redundant information continues to be compressed effectively. Then we propose the $\mathcal{L}_{ver}$ to enforce alignment between the outputs of original and quantized models. This calibration phase is critical, as it facilitates precise capture of each layer's output statistics—including activation ranges and distributions—which are employed to initialize quantization scaling factors. The overall process is shown in Algorithm~\ref{alg:ptq}.

%% file: sec/4_experiment.tex
\section{Experiments}
\label{sec:experiments}

\input{table/comp_dfd}

\input{table/comp_acc}
\input{table/supp_dfdcp}
\input{table/comp_compression}
\input{table/storage_compression}

\subsection{Setups}
\textbf{Dataset.} To validate the generalization ability of the proposed framework, we validate on \textbf{5} widely used Deepfake datasets: FaceForensics++ (FF++)~\cite{rossler2019faceforensics++}, DeepfakeDetection (DFD)~\cite{zi2020wilddeepfake}, Deepfake Detection Challenge (DFDC)~\cite{dolhansky2020deepfake}, the preview version of DFDC (DFDCP)~\cite{dolhansky2019deepfake}, and CelebDF (CDF)~\cite{li2020celeb}.
FF++ dataset is a widely used dataset in deepfake detection tasks, which consists of over 1.8 million forged images from 1,000 pristine videos. We use a subset of the training split of FF++ as the calibration dataset to calibrate and fine-tune the quantized model. The detailed procedure is elaborated in the Algorithm~\ref{alg:ptq}.




\textbf{Deepfake Detector.}
We test \emph{DeFakeQ}'s performance on \textbf{4} State-of-Art (SoTA) Detection methods, LSDA~\cite{yan2024transcending}, NPR~\cite{tan2024rethinking}, UCF~\cite{yan2023ucf}, and Wavelet~\cite{baru2025wavelet}, and \textbf{7} mainstream  backbones for model quantization,  DeiT (Tiny, Small, Base)~\cite{touvron2021training}, ViT (Small, Base)~\cite{dosovitskiy2020image}, and Swin (Small, Base)~\cite{liu2021swin}.

\textbf{Baselines.} To demonstrate that existing quantization methods has limited performance when directly applied to deepfake detection tasks, we compare with \textbf{3 State-of-Art general quantization baselines}, BRECQ~\cite{librecq}, Adalog~\cite{wu2024adalog}, and FIMA-Q~\cite{wu2025fima}. 
Despite of different techniques adopted, they are all general quantization methods. In addition, we have also compared with \textbf{5 SoTA compression methods in different technical categories}, including ADD~\cite{woo2022add}, GAC-FAS~\cite{le2024gradient}, X-Pruner~\cite{yu2023x}, One-bit~\cite{xu2024onebit}, HOLa~\cite{lei2025hola}.

\textbf{Evaluation Metrics.} We evaluate performance by detection accuracy (Area Under the Curve \textbf{AUC}) , storage cost (\textbf{MB}, deployment feasibility), and compression ratio (\textbf{\%}, size reduction relative to the full-precision model).

\textbf{Cross-dataset Evaluation Protocol.} Quantized models for each deepfake detector are trained on the FF++ dataset \cite{rossler2019faceforensics++} and evaluated on the CDF (v1, v2), DFD, DFDC, and DFDCP datasets.
Notice that the full-precision (FP) deepfake detection model, labeled ``GT'', serves as the upper bound for reference. 

\textbf{Implementation Details.} 
We employ \textbf{7} mainstream detectors as the original backbones for model quantization, encompassing DeiT (Tiny, Small, Base)~\cite{touvron2021training}, ViT (Small, Base)~\cite{dosovitskiy2020image}, and Swin (Small, Base)~\cite{liu2021swin}. All pre-trained detector parameters are initialized with pre-trained weights from the timm library. Following reconstruction-based PTQ approaches, 256 images are randomly sampled from the FF++ dataset to serve as the calibration set for quantizing deepfake detectors.
Consistent with established practices, our implemented quantization strategy utilizes per-channel asymmetric quantization for weights and per-tensor asymmetric quantization for activations. Regarding hyperparameter settings, $\alpha$, $\lambda$, $\gamma _{1}$, and $\gamma _{2}$ are configured as 1, 1, 0.5, and 0.5, respectively.

\subsection{Cross-Dataset Evaluation of Quantization Performance}
We validate our quantization performance through cross-dataset evaluations against SoTA quantization baselines, considering detection accuracy, compression rate, and storage cost. We further provide a visual analysis of attention maps to illustrate the underlying reasons for our superiority over existing SotA approaches.

\subsubsection{Accuracy.}
In summary, across all detection methods and across all datasets, \textbf{our \emph{DeFakeQ} substantially outperforms all quantization baselines on maintaining detection accuracy}. Table~\ref{tab:dfd}, Table~\ref{tab:acc} and Table~\ref{tab:dfdcp} present a comprehensive cross-data evaluation of quantization methods on Sota deepfake detections.
Specifically, as shown in Table~\ref{tab:dfd} , when evaluated on the DFD dataset, our method consistently outperforms \textbf{BRECQ}~\cite{librecq}, \textbf{Adalog}~\cite{wu2024adalog}, and \textbf{FIMA-Q}~\cite{wu2025fima} on both state-of-the-art and backbone detectors by \textbf{17.2\%--25.6\%}, \textbf{14.8\%--23.2\%}, and \textbf{13.5\%--25.7\%}, respectively. 
Furthermore, as summarized in Table~\ref{tab:acc} and Table~\ref{tab:dfdcp}, our approach maintains superior accuracy across multiple benchmarks. 
Compared with \textbf{BRECQ}, \textbf{Adalog}, and \textbf{FIMA-Q}, \emph{DeFakeQ} achieves improvements of \textbf{11.2\%--35.5\%}, \textbf{17.5\%--22.6\%}, and \textbf{11.2\%--23.7\%} on datasets \textbf{CDF-v1}, \textbf{CDF-v2}, \textbf{DFDC} and \textbf{DFDCP}, respectively, for both SOTA and backbone detectors, further demonstrating its robustness and generalization capability.
Notably, while achieving such state-of-the-art detection accuracy, our method remains comparable to existing quantization approaches in terms of compression rate and storage cost, showcasing its excellent balance between efficiency and performance for practical deepfake detection applications (refer to Table~\ref{tab:dfd}).

\subsubsection{Compression Rate and Storage Cost}
Table~\ref{tab:dfd} reports the storage cost of quantized models and their compression rates relative to the original full-precision detectors (GT). 
As shown in the table, \textbf{\emph{DeFakeQ} demonstrates competitive or superior efficiency in both storage cost and compression ratio, while, more importantly, achieving significantly improved detection accuracy.}
This performance gain can be attributed to our novel adaptive bidirectional quantization strategy, which effectively preserves critical forensic cues during the quantization process.

\subsubsection{Visual Analysis of Attention Maps} 
\begin{figure}[t]
\centering
\includegraphics[width=0.8\textwidth]{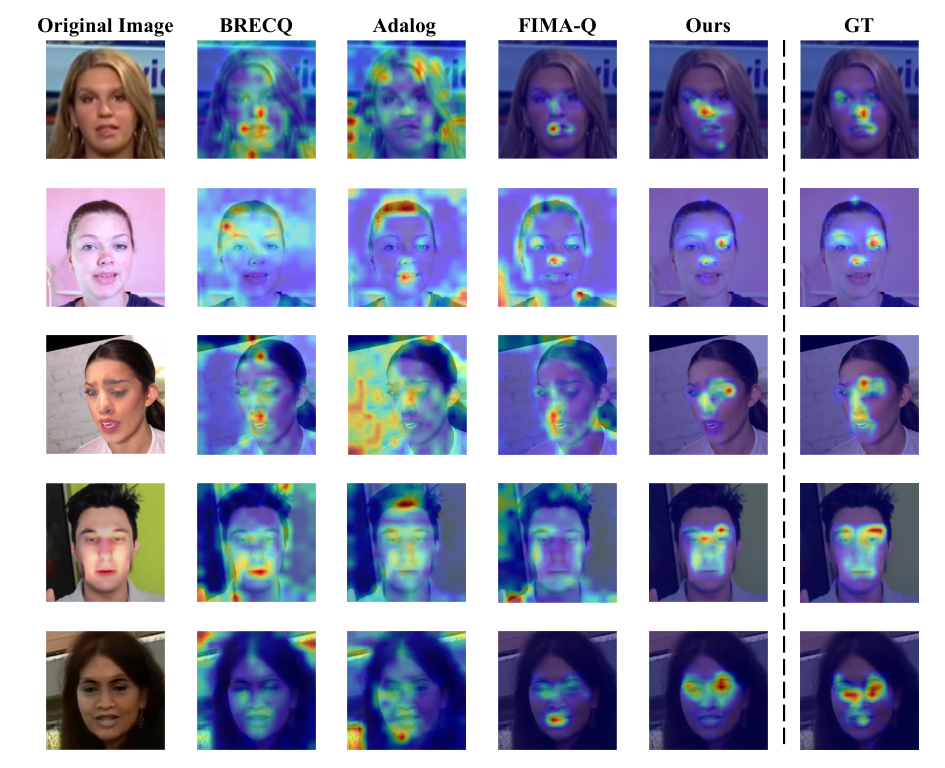}
\vspace{-5pt}
\caption{Visual Comparisons of detected salient forgeries during the quantization process.}
\label{fig:vis}
\vspace{-10pt}
\end{figure}

To analyze the performance of different quantization methods, we also visualize attention heatmaps. 
As shown in Figure~\ref{fig:vis}, BRECQ~\cite{librecq} and Adalog~\cite{wu2024adalog} exhibit significant discrepancies in the output of each attention block compared to the full-precision attention block (GT), which verifies that common quantization cannot be directly applied to deepfake detection tasks. Meanwhile, FIMA-Q~\cite{wu2025fima} alleviates this issue to a certain extent by employing the diagonal plus low-rank principle to preserve salient features, but it still has limitations for fine-grained deepfake detection tasks. In contrast, our \emph{DeFakeQ} can closely approximate the output of each layer in the GT attention block and accurately extract targeted feature regions (e.g., eyes, nose, mouth, etc.), thus greatly maintaining accuracy in deepfake detection during quantization process.

\subsection{Comparisons with Other Compression Methods}
 Table~\ref{quant_comparison} presents a comprehensive comparison of quantization results between our proposed method and other state-of-the-art model compression techniques across three benchmark datasets and three mainstream backbone models. Among the compared compression methods, ADD~\cite{woo2022add} (Knowledge Distillation), GAC-FAS~\cite{le2024gradient} (Neural architecture search), X-Pruner~\cite{yu2023x} (Model Pruning), One-bit~\cite{xu2024onebit} and HOLa~\cite{lei2025hola} (Low-rank Factorization) all exhibit varying degrees of performance degradation compared to GT, which is a common trade-off between model compression and performance preservation. In contrast, \textbf{our proposed \emph{DeFakeQ} method (based on quantization) maintains its superior performance advantage across all datasets and backbone models, consistently achieving the highest scores among all compressed methods}. Specifically, our \emph{DeFakeQ} achieves 74.5\% (DeiT-Base), 78.1\% (ViT-Base), and 77.1\% (Swin-Base) on the DFD dataset; 73.5\% (DeiT-Base), 76.4\% (ViT-Base), and 76.3\% (Swin-Base) on the CDF-v1 dataset; and 64.1\% (DeiT-Base), 67.2\% (ViT-Base), and 65.0\% (Swin-Base) on the DFDC dataset, respectively. These results fully validate the effectiveness of the quantization strategy adopted in \emph{DeFakeQ}, which can effectively mitigate performance loss while realizing model compression, outperforming other mainstream compression techniques.

Table \ref{storage_comparison} compares our DeFakeQ with other compression approaches in terms of storage cost and compression rate. The whole results are evaluated on the Swin-Base model and trained with the FF++ dataset~\cite{rossler2019faceforensics++}.
We observe that traditional compression strategies such as ADD~\cite{woo2022add}, GAC-FAS~\cite{le2024gradient}, and X-Pruner~\cite{yu2023x} achieve relatively high memory consumption and limited compression ratios, which restricts their deployment on resource-constrained edge platforms.
Although low-rank and binarization-based methods (One-bit~\cite{xu2024onebit}, HOLa~\cite{lei2025hola}) reduce memory usage to some extent, their accuracies still lag behind ours (refer to Table~\ref{quant_comparison}), which are unsuitable for edge devices for the real-world scenario.
In contrast, our DeFakeQ achieves the lowest storage cost and the highest compression rate, demonstrating the superior efficiency in model compression. These results validate that our quantization-aware design effectively minimizes memory overhead while maintaining strong compression performance, making it highly suitable for edge-oriented deepfake detection applications.

\subsection{Ablation Studies}

\subsubsection{Ablations on model design}
To validate the necessity and complementary effects of the proposed Horizontal Adaptive Quantization (HAQ) and Vertical Efficient Feature Fine-Tuning (VEFT), ablation experiments were conducted on five datasets. 

As shown in Table~\ref{tab:ablation}, removing either the HAQ or VEFT module leads to a significant drop in detection performance, thereby validating the contribution and effectiveness of each component.
The evaluated variants include the basic quantization, the baseline with the proposed Horizontal Adaptive Quantization (HAQ), the Vertical Efficienct Feature Fine-Tuning (VEFT), and our overall framework (HAQ + VEFT).
When only HAQ is enabled (\cmark \ HAQ, \xmark \ VEFT), the AUC scores improve significantly by 3.7–9.2\% compared to the baseline, demonstrating that HAQ’s adaptive quantization strategy effectively preserves critical feature information during model compression. Similarly, activating only VEFT (\xmark \ HAQ, \cmark \ VEFT) yields moderate improvements (5.0–11.4\% higher than the baseline), highlighting its ability to refine feature representations for better detection robustness. Furthermore, the full combination of HAQ and VEFT (Ours) attains the optimal performance across all datasets and backbones, outperforming the single-module implementations by 4.3-15.8\% and boosting the baseline by a large margin of 14.5-23.2\%. 

These results clearly illustrate that HAQ and VEFT play indispensable and synergistic roles: HAQ ensures efficient quantization with minimal information loss, while VEFT compensates for residual performance gaps by enhancing feature discriminability. Together, they enable the proposed method to achieve superior detection performance, underscoring the necessity of integrating both components for high-efficiency and high-accuracy deepfake detection.

\input{table/ablation}
\input{table/sensitivity_analysis}
\input{table/random_selection}

\subsubsection{Ablations Studies on Calibration Set Size}
We perform an experiment to demonstrate the effect of the calibration set size. Table~\ref{tab_calibration} presents the detection performance of \emph{DeFakeQ} with varying calibration set sizes on three representative deepfake detection benchmarks, evaluated using three mainstream transformer backbones.
From the empirical results, we observe that the detection accuracy rises steadily when the calibration set size is increased from 64 to 256, achieving the peak performance at 256 samples. Further expanding the calibration size to 512 and 1024 results in a mild degradation in accuracy across all datasets and backbone architectures, without yielding any observable performance improvement.
This observation clearly indicates that \emph{DeFakeQ} imposes only a low demand on the calibration set size. The underlying reason lies in that \textbf{\emph{DeFakeQ} is a quantization method} that aligns a quantized detector with a full-precision pre-trained model. Benefiting from this design principle, merely a small number of calibration samples are adequate to refine the calibration procedure, which effectively manifests the low-cost merit of our quantization framework in real-world deepfake detection applications.

\subsubsection{Ablation Studies on Random Selection Strategy in VEFT}
For each quantized block, VEFT randomly selects a small subset of feature channels to restore to the original full-precision format, consistent with the precision of the pre-trained full-precision Deepfake detector $M_f$. It is noted that the Random Selection is a data augmentation strategy, rather than a model design strategy, which is inspired by Dropout-style sample augmentation, and aims to focus the model on quantized channels by generating diverse mixed-precision samples in a self-supervised manner. With the design of progressive contrastive learning, VEFT adaptively optimizes the quantized channels. To explore the influence of the random selection strategy on the detection performance, ablation experiments were conducted on five datasets, which is shown in Table~\ref{tab:random}.
The three compared methods are designed to clarify the effect of the channel selection strategy when adopting VEFT. Specifically, the first method corresponds to the scenario without adopting VEFT, which is marked as "\xmark \ VEFT" in the table. It can be observed that this method achieves the lowest detection accuracy across all datasets and backbones, indicating that the absence of VEFT significantly limits the model performance.

The second method adopts VEFT but adopts a fixed selection strategy, i.e., adopting VEFT while fixedly selecting a small subset of feature channels to restore to the original full-precision format, denoted as "\cmark \ VEFT (\xmark \ Random Selection)". Compared with the first method without VEFT, this method achieves a noticeable performance improvement on all datasets and backbones. This improvement demonstrates that restoring a small subset of feature channels to full precision under VEFT can effectively enhance the detection capability of the model.

The third method is the proposed strategy, which involves adopting VEFT and randomly selecting feature channels to restore to full precision, labeled as "\cmark \ VEFT (\cmark \ Random Selection)". As shown in the table, this method achieves the highest detection accuracy across all datasets and backbones, outperforming the other two methods consistently. This result fully verifies that the random channel selection strategy under VEFT can better exploit the feature information, thereby further improving the deepfake detection performance.

\subsection{Real-Time, Real-World Deployment on Edge Devices}
\begin{figure}[t]
\centering
\includegraphics[width=0.8\textwidth]{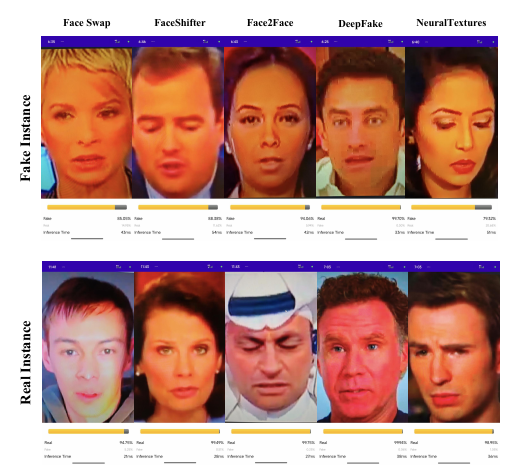}
\caption{Real-world detection instances of \emph{DeFakeQ} deployed on an Android device. The quantized model processes in around 35 ms, and outputs the prediction result and the corresponding confidence score. Please refer to the demo video for our real time and real work depoyment on the android phone.}
\label{fig:edgeins}
\vspace{-10pt}
\end{figure}

To validate the practical applicability of \emph{DeFakeQ}, we deploy it on Android mobile phones and conducted edge-device testing on five prominent face forgery methods, including Face Swap~\cite{zhu2021one}, Face Shifter~\cite{li2020advancing}, Face2Face~\cite{thies2016face2face}, DeepFake~\cite{yan2024df40}, and NeuralTextures~\cite{thies2019deferred}, as shown in Figure~\ref{fig:edgeins}. 

In addition, we have recorded a video to showcase the practical deployment and performance of our \emph{DeFakeQ} on edge mobile devices. 
\textbf{Please refer to the demo video for our real time and real work depoyment on the android phone.}

The demo validates that \emph{DeFakeQ} has been successfully installed and runs seamlessly on mobile hardware, enabling reliable differentiation between real and fake images. Notably, the fake images in the demo cover multiple common deepfake categories, e.g., face swapping, face2face, and neuraltexture, and our method consistently achieves accurate detection results across all these scenarios. Equally impressive is the real-time inference capability of \emph{DeFakeQ}: the inference latency per image is no more than 50ms, which fully meets the requirements for real-time applications. This demonstration not only confirms the compatibility and feasibility of \emph{DeFakeQ} on edge mobile devices but also highlights its superior balance between detection accuracy and efficiency, making it a promising solution for real-world deepfake detection tasks.

During real-time inference, \emph{DeFakeQ} achieves an average latency of 25 ms per frame and consumes 87 mW of power while maintaining high detection accuracy. Our \emph{DeFakeQ} performance balances efficiency and reliability, enabling seamless integration into mobile deepfake detection apps for end-users to verify media authenticity. 

\textbf{Discussions for broader real-world adoptions}\quad 
\emph{DeFakeQ}  can be further adapted to custom circuit boards (e.g., ASICs and FPGAs) for edge deployment scenarios. The proposed HAQ module can be aligned with the low-bit arithmetic capabilities of modern circuit designs. Besides, our \emph{DeFakeQ} is hardware agnostic and adheres to standard integer arithmetic specifications. HAQ and VEFT are 
compatible with iOS via Core ML, embedded systems, e.g., Raspberry Pi, ESP32, and custom circuit boards like ASICs/FPGAs. In addition, \emph{DeFakeQ}’s output can be directly exported to ONNX for cross-platform deployment or PyTorch Mobile for Python-based edge applications.
Leveraging the ultra-low storage cost (as verified in Table~\ref{tab:dfd}) and optimized computational graph, \emph{DeFakeQ} occupies far below the threshold for lightweight mobile applications.

\subsection{Limitations and Future Works}
Current real-world deployment is limited to mobile devices; we have not extended the validation to more resource-constrained edge scenarios, e.g., low-power IoT devices, edge gateways, or evaluated performance under extreme environmental conditions, e.g., low-light imaging, partial facial occlusion. 
We plan to extend deployment to a broader range of edge platforms and optimize a resource-aware dynamic quantization module that adjusts precision in real time based on device residual computing power and network bandwidth.


%% file: table/comp_dfd.tex
\begin{table*}[t]
\caption{\textbf{Cross-dataset evaluations} of Detection Accuracy, Storage Cost and Compression Rate on testing Dataset DFD~\cite{zi2020wilddeepfake}. 
All quantized detectors are trained on FF++~\cite{rossler2019faceforensics++}. 
\textbf{Our method achieves significantly improved detection accuracy (\%) comparable to existing quantization approaches}, while remains comparable in Compression Rate (\%) and Storage Cost (MB).}
\label{tab:dfd}
\centering
\scriptsize
\renewcommand{\arraystretch}{1.2}
\setlength{\tabcolsep}{1.2pt}

\begin{tabular}{lc|ccccc|cccc|ccccc}
\toprule
\textbf{Detectors} & \textbf{Type} &
\multicolumn{5}{c|}{\textbf{Storage Cost (MB) $\downarrow$}} &
\multicolumn{4}{c|}{\textbf{Compression Rate (\%) $\uparrow$}} &
\multicolumn{5}{c}{\textbf{Detection Accuracy (\%) $\uparrow$}} \\ 
\cmidrule(lr){3-7} \cmidrule(lr){8-11} \cmidrule(lr){12-16}
& &
\textbf{\shortstack{BRECQ\\ICLR'22}} &
\textbf{\shortstack{Adalog\\ECCV'24}} &
\textbf{\shortstack{FIMA-Q\\CVPR'25}} &
\textbf{Ours} & \textbf{GT} &
\textbf{\shortstack{BRECQ\\ICLR'22}} &
\textbf{\shortstack{Adalog\\ECCV'24}} &
\textbf{\shortstack{FIMA-Q\\CVPR'25}} &
\textbf{Ours} &
\textbf{\shortstack{BRECQ\\ICLR'22}} &
\textbf{\shortstack{Adalog\\ECCV'24}} &
\textbf{\shortstack{FIMA-Q\\CVPR'25}} &
\textbf{Ours} & \textbf{GT} \\ 
\midrule
DeiT-Tiny    & \multirow{7}{*}{Backbones} 
& 5.9  & 2.8  & 2.4  & 2.2  & 23.7 
& 75.0 & 88.2 & 90.0 & 90.7 
& 52.1 (-17.2\%) & 54.5 (-14.8\%) & 53.2 (-16.1\%) & \textbf{69.3} & 77.8 \\
DeiT-Small   & 
& 20.2 & 9.3  & 7.6  & 7.5  & 84.1 
& 76.0 & 88.9 & 91.0 & 91.1 
& 50.3 (-21.8\%) & 52.8 (-19.3\%) & 54.1 (-18.0\%) & \textbf{72.1} & 79.4 \\
DeiT-Base    & 
& 75.9 & 33.0 & 26.4 & 31.7 & 330.2 
& 77.0 & 90.0 & 92.0 & 90.4 
& 58.6 (-15.9\%) & 56.9 (-17.6\%) & 60.3 (-14.2\%) & \textbf{74.5} & 82.3 \\
ViT-Small    & 
& 21.0 & 10.1 & 8.4  & 8.2  & 84.1 
& 75.0 & 88.0 & 90.0 & 90.2 
& 56.4 (-20.9\%) & 58.2 (-19.1\%) & 61.7 (-15.6\%) & \textbf{77.3} & 84.5 \\
ViT-Base     & 
& 79.2 & 29.7 & 26.4 & 31.2 & 330.2 
& 76.0 & 91.0 & 92.0 & 90.6 
& 57.2 (-20.9\%) & 58.8 (-19.3\%) & 61.1 (-17.0\%) & \textbf{78.1} & 86.7 \\
Swin-Small   & 
& 43.5 & 20.8 & 17.0 & 18.1 & 189.2 
& 77.0 & 88.9 & 91.0 & 90.4 
& 55.9 (-21.0\%) & 53.7 (-23.2\%) & 63.4 (-13.5\%) & \textbf{76.9} & 85.0 \\
Swin-Base    & 
& 73.7 & 33.5 & 26.8 & 31.8 & 334.8 
& 78.0 & 90.0 & 92.0 & 90.5 
& 51.5 (-25.6\%) & 55.8 (-21.3\%) & 60.9 (-16.2\%) & \textbf{77.1} & 86.5 \\
\midrule[0.8pt]
LSDA         & CVPR'24 
& 58.2 & 27.9 & 22.8 & 26.6 & 253.2 
& 77.0 & 88.9 & 91.0 & 89.5 
& 52.8 (-25.6\%) & 56.3 (-22.1\%) & 58.1 (-20.3\%) & \textbf{78.4} & 88.0 \\
NPR          & CVPR'24 
& 42.7 & 21.4 & 17.8 & 19.7 & 177.9 
& 76.0 & 88.0 & 90.0 & 88.9 
& 50.4 (-27.2\%) & 55.2 (-22.4\%) & 56.7 (-20.9\%) & \textbf{77.6} & 85.5 \\
UCF          & ICCV'23 
& 22.9 & 11.9 & 10.1 & 9.1  & 91.6  
& 75.0 & 87.0 & 89.0 & 90.0 
& 43.7 (-33.6\%) & 48.2 (-29.1\%) & 50.1 (-27.2\%) & \textbf{77.3} & 84.9 \\
Wavelet      & WACV'25 
& 135.3& 61.5 & 49.2 & 84.8 & 615.0 
& 78.0 & 90.0 & 92.0 & 86.2 
& 43.9 (-35.0\%) & 47.8 (-31.1\%) & 51.2 (-27.7\%) & \textbf{78.9} & 85.7 \\ 
\bottomrule
\end{tabular}
\end{table*}

%% file: table/comp_acc.tex
\begin{table*}[t]
\caption{Additional \textbf{Cross-datasets evaluation} of Detection Accuracy
on testing datasets CDF-v1~\cite{li2020celeb}, CDF-v2~\cite{li2020celeb}, and DFDC~\cite{dolhansky2019deepfake}.
All quantized detectors are trained on the FF++ dataset~\cite{rossler2019faceforensics++}, with the best results highlighted in bold.}
\label{tab:acc}
\centering
\tiny
\renewcommand{\arraystretch}{1.5}
\setlength{\tabcolsep}{1.2pt}

\begin{tabular}{lc|ccccc|ccccc|ccccc}
\toprule
\textbf{Detectors} & \textbf{Type} &
\multicolumn{5}{c|}{\textbf{CDF-v1 (Accuracy\%)}} &
\multicolumn{5}{c|}{\textbf{CDF-v2 (Accuracy\%)}} &
\multicolumn{5}{c}{\textbf{DFDC (Accuracy\%)}} \\
\cmidrule(lr){3-7} \cmidrule(lr){8-11} \cmidrule(lr){12-16}
& &
\textbf{\shortstack{BRECQ\\ICLR'22}} &
\textbf{\shortstack{Adalog\\ECCV'24}} &
\textbf{\shortstack{FIMA-Q\\CVPR'25}} &
\textbf{Ours} & \textbf{GT} &
\textbf{\shortstack{BRECQ\\ICLR'22}} &
\textbf{\shortstack{Adalog\\ECCV'24}} &
\textbf{\shortstack{FIMA-Q\\CVPR'25}} &
\textbf{Ours} & \textbf{GT} &
\textbf{\shortstack{BRECQ\\ICLR'22}} &
\textbf{\shortstack{Adalog\\ECCV'24}} &
\textbf{\shortstack{FIMA-Q\\CVPR'25}} &
\textbf{Ours} & \textbf{GT} \\
\midrule
DeiT-Tiny    & \multirow{7}{*}{Backbones}
& 50.2 (-18.3\%) & 55.8 (-12.7\%) & 54.3 (-14.2\%) & \textbf{68.5} & 76.7
& 45.3 (-19.2\%) & 47.6 (-16.9\%) & 49.9 (-14.6\%) & \textbf{64.5} & 73.3
& 37.2 (-20.5\%) & 39.4 (-18.3\%) & 40.3 (-17.4\%) & \textbf{57.7} & 65.1 \\
DeiT-Small   &
& 48.9 (-22.2\%) & 51.9 (-19.2\%) & 55.0 (-16.1\%) & \textbf{71.1} & 78.2
& 42.1 (-25.9\%) & 50.1 (-17.9\%) & 50.5 (-17.5\%) & \textbf{68.0} & 74.8
& 36.1 (-23.9\%) & 40.0 (-20.0\%) & 42.5 (-17.5\%) & \textbf{60.0} & 66.4 \\
DeiT-Base    &
& 60.8 (-12.7\%) & 58.5 (-15.0\%) & 61.2 (-12.3\%) & \textbf{73.5} & 81.2
& 46.6 (-22.7\%) & 48.2 (-21.1\%) & 49.8 (-19.5\%) & \textbf{69.3} & 77.8
& 42.7 (-21.4\%) & 41.6 (-22.5\%) & 47.5 (-16.6\%) & \textbf{64.1} & 71.4 \\
ViT-Small    &
& 58.1 (-17.9\%) & 59.7 (-16.3\%) & 62.8 (-13.2\%) & \textbf{76.0} & 82.0
& 44.7 (-29.3\%) & 46.6 (-27.4\%) & 52.2 (-21.8\%) & \textbf{74.0} & 79.8
& 43.1 (-23.7\%) & 42.8 (-24.0\%) & 47.1 (-19.7\%) & \textbf{66.8} & 72.1 \\
ViT-Base     &
& 58.9 (-17.5\%) & 60.0 (-16.4\%) & 62.2 (-14.2\%) & \textbf{76.4} & 83.7
& 48.2 (-26.5\%) & 47.2 (-27.5\%) & 56.3 (-18.4\%) & \textbf{74.7} & 81.9
& 45.2 (-22.0\%) & 45.1 (-22.1\%) & 52.4 (-14.8\%) & \textbf{67.2} & 73.6 \\
Swin-Small   &
& 57.7 (-18.4\%) & 55.3 (-20.8\%) & 64.9 (-11.2\%) & \textbf{76.1} & 83.8
& 45.5 (-26.7\%) & 43.9 (-28.3\%) & 54.5 (-17.7\%) & \textbf{72.2} & 80.2
& 45.9 (-18.7\%) & 43.3 (-21.3\%) & 53.4 (-11.2\%) & \textbf{64.6} & 71.1 \\
Swin-Base    &
& 53.2 (-23.1\%) & 57.5 (-18.8\%) & 62.0 (-14.3\%) & \textbf{76.3} & 85.8
& 48.0 (-25.7\%) & 45.0 (-28.7\%) & 50.1 (-23.6\%) & \textbf{73.7} & 82.9
& 41.5 (-14.5\%) & 43.5 (-12.5\%) & 50.9 (-5.1\%) & \textbf{56.0} & 73.2 \\
\midrule[0.8pt]
LSDA         & CVPR'24
& 54.0 (-24.9\%) & 58.0 (-20.9\%) & 59.8 (-19.1\%) & \textbf{78.9} & 86.7
& 46.6 (-28.7\%) & 49.2 (-26.1\%) & 52.0 (-23.3\%) & \textbf{75.3} & 83.0
& 42.8 (-23.1\%) & 46.9 (-19.0\%) & 49.8 (-16.1\%) & \textbf{65.9} & 73.6 \\
NPR          & CVPR'24
& 52.1 (-26.2\%) & 56.9 (-21.4\%) & 57.4 (-20.9\%) & \textbf{78.3} & 85.8
& 46.3 (-27.9\%) & 48.7 (-25.5\%) & 49.7 (-24.5\%) & \textbf{74.2} & 81.2
& 40.4 (-25.7\%) & 42.5 (-23.6\%) & 51.6 (-14.5\%) & \textbf{66.1} & 75.8 \\
UCF          & ICCV'23
& 45.3 (-25.4\%) & 49.8 (-20.9\%) & 51.7 (-19.0\%) & \textbf{70.7} & 77.9
& 46.1 (-23.2\%) & 42.1 (-27.2\%) & 47.9 (-21.4\%) & \textbf{69.3} & 75.3
& 39.7 (-27.6\%) & 42.8 (-24.5\%) & 55.5 (-11.8\%) & \textbf{67.3} & 74.0 \\
Wavelet      & WACV'25
& 45.1 (-24.5\%) & 50.0 (-19.6\%) & 52.5 (-17.1\%) & \textbf{69.6} & 75.6
& 48.5 (-22.4\%) & 43.6 (-27.3\%) & 50.6 (-20.3\%) & \textbf{70.9} & 75.9
& 46.6 (-22.3\%) & 43.9 (-25.0\%) & 53.1 (-15.8\%) & \textbf{68.9} & 75.3 \\
\bottomrule
\end{tabular}
\end{table*}

%% file: table/supp_dfdcp.tex
\begin{table*}[htb]
\caption{Additional \textbf{Cross-datasets evaluation} of Detection Accuracy on testing dataset DFDCP~\cite{dolhansky2019deepfake} where all quantized detectors are trained on FF++~\cite{rossler2019faceforensics++}. 
}
\label{tab:dfdcp}
\centering
\scriptsize
\renewcommand{\arraystretch}{1.0}
\setlength{\tabcolsep}{8.0pt}

\begin{tabular}{lc|ccccc}
\toprule
 &
   &
  \multicolumn{5}{c}{\textbf{Dateset DFDCP~\cite{dolhansky2019deepfake} (Accuracy\%)}} \\ \cline{3-7} 
 &
   &
  \multicolumn{4}{c:}{\textbf{Quantization Methods}} &
  \textbf{} \\ \cline{3-6}
\multirow{-3}{*}{\textbf{Detectors}} &
  \multirow{-3}{*}{\textbf{Type}} &
  \textbf{\begin{tabular}[c]{@{}c@{}}BRECQ~\cite{librecq}\\ ICLR'22\end{tabular}} &
  \textbf{\begin{tabular}[c]{@{}c@{}}Adalog~\cite{wu2024adalog}\\ ECCV'24\end{tabular}} &
  \textbf{\begin{tabular}[c]{@{}c@{}}FIMA-Q~\cite{wu2025fima}\\ CVPR'25\end{tabular}} &
  \multicolumn{1}{c:}{\textbf{Ours}} &
  {\color[HTML]{656565} \textbf{GT}} \\ 
  \midrule
DeiT-Tiny &
  \multirow{7}{*}{Backbones} &
  43.3(-20.4\%) &
  44.7(-18.8\%) &
  54.1(-15.1\%) &
  \multicolumn{1}{c:}{\textbf{63.7}} &
  {\color[HTML]{656565} 72.1} \\
DeiT-Small &
  &
  44.1(-33.6\%) &
  52.9(-20.3\%) &
  55.1(-17.0\%) &
  \multicolumn{1}{c:}{\textbf{66.4}} &
  {\color[HTML]{656565} 73.5} \\
DeiT-Base &
  &
  46.1(-33.4\%) &
  55.3(-20.1\%) &
  57.6(-16.8\%) &
  \multicolumn{1}{c:}{\textbf{69.2}} &
  {\color[HTML]{656565} 76.8} \\
ViT-Small &
  &
  46.2(-35.5\%) &
  55.4(-22.6\%) &
  57.8(-19.3\%) &
  \multicolumn{1}{c:}{\textbf{71.6}} &
  {\color[HTML]{656565} 77.0} \\
ViT-Base &
  &
  48.1(-33.5\%) &
  57.7(-20.2\%) &
  60.2(-16.7\%) &
  \multicolumn{1}{c:}{\textbf{72.3}} &
  {\color[HTML]{656565} 80.2} \\
Swin-Small &
  &
  47.2(-33.0\%) &
  56.7(-19.6\%) &
  59.0(-16.3\%) &
  \multicolumn{1}{c:}{\textbf{70.5}} &
  {\color[HTML]{656565} 78.7} \\
Swin-Base &
  &
  48.5(-32.5\%) &
  58.2(-19.1\%) &
  60.7(-15.6\%) &
  \multicolumn{1}{c:}{\textbf{71.9}} &
  {\color[HTML]{656565} 80.9} \\
\midrule[0.8pt]
LSDA~\cite{yan2024transcending} &
  CVPR'24 &
  48.8(-31.8\%) &
  58.5(-18.2\%) &
  61.0(-14.7\%) &
  \multicolumn{1}{c:}{\textbf{71.5}} &
  {\color[HTML]{656565} 81.3} \\
NPR~\cite{tan2024rethinking} &
  CVPR'24 &
  49.2(-31.8\%) &
  59.0(-18.2\%) &
  61.5(-14.7\%) &
  \multicolumn{1}{c:}{\textbf{72.1}} &
  {\color[HTML]{656565} 82.0} \\
UCF~\cite{yan2023ucf} &
  ICCV'23 &
  47.1(-32.9\%) &
  56.5(-19.5\%) &
  58.9(-16.1\%) &
  \multicolumn{1}{c:}{\textbf{70.2}} &
  {\color[HTML]{656565} 78.5} \\
Wavelet~\cite{baru2025wavelet} &
  WACV'25 &
  42.1(-41.6\%) &
  50.5(-30.0\%) &
  52.6(-27.0\%) &
  \multicolumn{1}{c:}{\textbf{72.1}} &
  {\color[HTML]{656565} 70.1} \\ 
  \bottomrule
\end{tabular}

\end{table*}

%% file: table/comp_compression.tex
\begin{table*}[t]
\centering
\caption{Comparison with compression methods across other technical categories on detection accuracy(\%).}
\label{quant_comparison}
\scriptsize
\renewcommand{\arraystretch}{1.2}
\setlength{\tabcolsep}{0.75pt}
\setlength{\aboverulesep}{0pt}
\setlength{\belowrulesep}{0pt}
\begin{tabular}{lcccccccccccc}
\toprule
\textbf{Methods} & \textbf{Venue} & \textbf{Technology Category} & \multicolumn{3}{c}{\textbf{DFD}~\cite{zi2020wilddeepfake}} & \multicolumn{3}{c}{\textbf{CDF-v1}~\cite{li2020celeb}} & \multicolumn{3}{c}{\textbf{DFDC}~\cite{dolhansky2020deepfake}} \\
\cmidrule(lr){4-6} \cmidrule(lr){7-9} \cmidrule(l){10-12}
& & & DeiT-Base & ViT-Base & Swin-Base & DeiT-Base & ViT-Base & Swin-Base & DeiT-Base & ViT-Base & Swin-Base \\
\midrule

ADD~\cite{woo2022add}           & AAAI'22     & Knowledge Distillation         & 49.5 & 46.6 & 49.0 & 47.7 & 48.2 & 49.0 & 37.1 & 36.8 & 35.2 \\
GAC-FAS~\cite{le2024gradient}       & CVPR'24     & Neural Architecture Search        & 52.1 & 51.3 & 50.6 & 45.1 & 48.9 & 47.6 & 45.1 & 45.0 & 41.3 \\
X-Pruner~\cite{yu2023x}      & CVPR'23     & Model Pruning    & 52.1 & 55.5 & 59.1 & 53.9 & 51.0 & 51.5 & 48.6 & 47.2 & 50.1 \\
One-bit~\cite{xu2024onebit}       & NIPS'24     & Low-Rank Factorization   & 56.6 & 53.2 & 52.5 & 50.0 & 48.1 & 47.1 & 40.5 & 43.2 & 42.7 \\
HOLa~\cite{lei2025hola}       & ICCV'25     & Low-Rank Factorization   & 59.2 & 57.0 & 55.4 & 48.8 & 51.9 & 49.9 & 46.5 & 50.2 & 52.7 \\
\textbf{\emph{DeFakeQ} (Ours)}  & -             & Quantization     & \textbf{74.5} & \textbf{78.1} & \textbf{77.1} & \textbf{73.5} & \textbf{76.4} & \textbf{76.3} & \textbf{64.1} & \textbf{67.2} & \textbf{65.0} \\
\hdashline
Ground Truth (GT)            & -             & -                    & 82.3 & 86.7 & 86.5 & 81.2 & 83.7 & 85.8 & 71.4 & 73.6 & 73.2 \\
\bottomrule
\end{tabular}
\end{table*}

%% file: table/storage_compression.tex
\begin{table}[t]
\centering
\caption{Comparison with compression methods across other technical categories on Storage Cost (MB) and Compression Rate (\%). $\downarrow$ means the lower the better and $\uparrow$ means the higher the better.}
\label{storage_comparison}
\scriptsize
\renewcommand{\arraystretch}{1.1}
\setlength{\tabcolsep}{50pt}
\setlength{\aboverulesep}{0pt}
\setlength{\belowrulesep}{0pt}
\resizebox{1\columnwidth}{!}{%
\begin{tabular}{lcc}
\toprule
\textbf{Methods} 
& \textbf{Storage Cost (MB)} $\downarrow$ 
& \textbf{Compression Rate (\%)} $\uparrow$ \\
\midrule

ADD~\cite{woo2022add}           & 230.2 & 31.2 \\
GAC-FAS~\cite{le2024gradient}   & 150.7 & 55.0 \\
X-Pruner~\cite{yu2023x}         & 133.9 & 60.0 \\
One-bit~\cite{xu2024onebit}     & 72.7  & 78.3 \\
HOLa~\cite{lei2025hola}         & 86.6  & 74.1 \\
\textbf{\emph{DeFakeQ} (Ours)}  & \textbf{31.8} & \textbf{90.5} \\
\hdashline
Ground Truth (GT)               & 334.8 & - \\
\bottomrule
\end{tabular}
}
\vspace{-5pt}
\end{table}

%% file: table/ablation.tex
\begin{table*}[t]
\centering
\caption{Ablation studies on model design (HAQ and VEFT). 
All detection models are trained on the FF++ dataset~\cite{rossler2019faceforensics++} and evaluated across various other datasets on metric detection accuracy(\%). 
}
\label{tab:ablation}
\tiny
\renewcommand{\arraystretch}{1.2}
\setlength{\tabcolsep}{1.0pt}
\begin{tabular}{lccccccccccccccc}
\toprule[1.2pt]
\multicolumn{1}{c}{\multirow{2}{*}{\textbf{Method}}} 
& \multicolumn{3}{c}{\textbf{CDF-v1}~\cite{li2020celeb}} 
& \multicolumn{3}{c}{\textbf{CDF-v2}~\cite{li2020celeb}} 
& \multicolumn{3}{c}{\textbf{DFD}~\cite{zi2020wilddeepfake}} 
& \multicolumn{3}{c}{\textbf{DFDC}~\cite{dolhansky2020deepfake}}
& \multicolumn{3}{c}{\textbf{DFDCP}~\cite{dolhansky2019deepfake}} \\
\cmidrule(lr){2-4} \cmidrule(lr){5-7} \cmidrule(lr){8-10} \cmidrule(lr){11-13} \cmidrule(lr){14-16}
& DeiT-Base & ViT-Base & Swin-Base 
& DeiT-Base & ViT-Base & Swin-Base 
& DeiT-Base & ViT-Base & Swin-Base 
& DeiT-Base & ViT-Base & Swin-Base
& DeiT-Base & ViT-Base & Swin-Base \\
\midrule[0.8pt]
\xmark  \ HAQ, \xmark  \ VEFT                 & 57.8 & 58.9 & 53.2 & 46.6 & 48.2 & 48.0 & 52.2 & 54.9 & 51.9 & 42.3 & 45.8 & 43.4 & 52.5 & 57.1 & 55.8 \\
\cmark \ HAQ, \xmark \ VEFT & 61.5 & 63.3 & 60.7 & 58.4 & 59.1 & 58.2 & 59.7 & 62.4 & 60.0 & 50.1 & 53.5 & 56.0 & 60.4 & 62.5 & 61.7 \\
\xmark \ HAQ, \cmark \ VEFT  & 65.2 & 61.9 & 66.6 & 64.5 & 60.8 & 68.2 & 63.6 & 64.4 & 62.4 & 55.9 & 57.8 & 60.1 & 57.2 & 59.8 & 60.2 \\
\midrule[0.8pt]
\textbf{\cmark \ HAQ, \cmark \ VEFT (Ours)}        & \textbf{73.5} & \textbf{76.4} & \textbf{76.3} & \textbf{69.3} & \textbf{74.7} & \textbf{73.7} & \textbf{74.5} & \textbf{78.1} & \textbf{77.1} & \textbf{64.1} & \textbf{67.2} & \textbf{65.0} & \textbf{69.2} & \textbf{72.3} & \textbf{71.9} \\
\bottomrule[1.2pt]
\end{tabular}
\vspace{-10pt}
\end{table*}

%% file: table/sensitivity_analysis.tex
\begin{table*}[htb]
\centering
\renewcommand{\arraystretch}{1.1}
\setlength{\tabcolsep}{12pt}
\caption{Ablations Studies on Calibration Set Size. All detection models are trained on the FF++ dataset~\cite{rossler2019faceforensics++} and evaluated across various other datasets on metric detection accuracy(\%).
}
\label{tab_calibration}
\small
\resizebox{1\textwidth}{!}{
\begin{tabular}{cccccccccc}
\toprule[0.8pt]
\multirow{2}{*}{\textbf{Calibration Set Size}} & \multicolumn{3}{c}{\textbf{DFD}~\cite{zi2020wilddeepfake}} & \multicolumn{3}{c}{\textbf{CDF-v1}~\cite{li2020celeb}} & \multicolumn{3}{c}{\textbf{DFDC}~\cite{dolhansky2020deepfake}} \\
\cmidrule(lr){2-4} \cmidrule(lr){5-7} \cmidrule(l){8-10}
& \textbf{DeiT-Base} & \textbf{ViT-Base} & \textbf{Swin-Base} & \textbf{DeiT-Base} & \textbf{ViT-Base} & \textbf{Swin-Base} & \textbf{DeiT-Base} & \textbf{ViT-Base} & \textbf{Swin-Base} \\
\midrule[0.5pt]
64  & 69.8 & 72.6 & 71.3 & 71.5 & 74.8 & 72.9 & 61.3 & 61.2 & 62.7 \\
128 & 72.1 & 75.8 & 74.7 & 72.2 & 75.0 & 75.5 & 63.1 & 64.5 & 63.3 \\
\textbf{256} & \textbf{74.5} & \textbf{78.1} & \textbf{77.1} & \textbf{73.5} & \textbf{76.4} & \textbf{76.3} & \textbf{64.1} & \textbf{67.2} & \textbf{65.0} \\
512 & 73.4 & 77.0 & 75.8 & 72.9 & 75.8 & 75.8 & 63.8 & 65.7 & 64.2 \\
1024& 73.1 & 76.8 & 75.5 & 72.6 & 75.5 & 75.6 & 63.5 & 65.3 & 64.0 \\
\bottomrule[0.8pt]
\end{tabular}
}
\vspace{-8pt}
\end{table*}

%% file: table/random_selection.tex
\begin{table*}[t]
\centering
\caption{Ablations Studies on Random Selection Strategy. All detection models are trained on the FF++ dataset~\cite{rossler2019faceforensics++} and evaluated across various other datasets on metric detection accuracy(\%).
}
\label{tab:random}
\tiny
\renewcommand{\arraystretch}{1.4}
\setlength{\tabcolsep}{0.4pt}
\begin{tabular}{lccccccccccccccc}
\toprule[1.2pt]
\multicolumn{1}{c}{\multirow{2}{*}{\textbf{Method}}} 
& \multicolumn{3}{c}{\textbf{CDF-v1}~\cite{li2020celeb}} 
& \multicolumn{3}{c}{\textbf{CDF-v2}~\cite{li2020celeb}} 
& \multicolumn{3}{c}{\textbf{DFD}~\cite{zi2020wilddeepfake}} 
& \multicolumn{3}{c}{\textbf{DFDC}~\cite{dolhansky2020deepfake}}
& \multicolumn{3}{c}{\textbf{DFDCP}~\cite{dolhansky2019deepfake}} \\
\cmidrule(lr){2-4} \cmidrule(lr){5-7} \cmidrule(lr){8-10} \cmidrule(lr){11-13} \cmidrule(lr){14-16}
& DeiT-Base & ViT-Base & Swin-Base 
& DeiT-Base & ViT-Base & Swin-Base 
& DeiT-Base & ViT-Base & Swin-Base 
& DeiT-Base & ViT-Base & Swin-Base
& DeiT-Base & ViT-Base & Swin-Base \\
\midrule[0.8pt]
 \xmark  \ VEFT                 & 61.5 & 63.3 & 60.7 & 58.4 & 59.1 & 58.2 & 59.7 & 62.4 & 60.0 & 50.1 & 53.5 & 56.0 & 60.4 & 62.5 & 61.7 \\
 \cmark \ VEFT (\xmark \ Random Selection)  & 70.2 & 69.1 & 72.3 & 65.9 & 70.8 & 69.2 & 67.6 & 71.5 & 72.3 & 61.5 & 62.6 & 61.8 & 64.4 & 68.9 & 67.2 \\
\midrule[0.8pt]
\textbf{\cmark \ VEFT (\cmark \ Random Selection)}        & \textbf{73.5} & \textbf{76.4} & \textbf{76.3} & \textbf{69.3} & \textbf{74.7} & \textbf{73.7} & \textbf{74.5} & \textbf{78.1} & \textbf{77.1} & \textbf{64.1} & \textbf{67.2} & \textbf{65.0} & \textbf{69.2} & \textbf{72.3} & \textbf{71.9} \\
\bottomrule[1.2pt]
\end{tabular}
\vspace{-10pt}
\end{table*}

%% file: sec/5_conclusion.tex
\section{Conclusion}
\label{sec:conclusion}
To conclude, we propose \emph{DeFakeQ}, the first specialized quantization framework tailored for deepfake detection, addressing the challenge of deploying high-performing detectors on resource-constrained edge devices. Unlike traditional quantization methods that tend to disrupt the subtle facial forgery cues critical to reliable detection, \emph{DeFakeQ} introduces a novel adaptive bidirectional quantization, which achieves substantial model compression while preserving the discriminative capacity to capture fine-grained facial forgery cues. Extensive evaluations demonstrate its superiority over conventional quantization methods in maintaining post-quantization accuracy, while real-world deployment on edge platforms further substantiates its practicality and robustness. Furthermore, real-world deployment on mobile devices further confirms \emph{DeFakeQ}’s practicality, these results underscore the potential of \emph{DeFakeQ} as a viable solution for real-time, on-device deepfake detection in practical applications.